\documentclass{article}
\usepackage{ijcai18}

\usepackage{times}
\usepackage{xcolor}
\usepackage{soul}
\usepackage[utf8]{inputenc}
\usepackage[small]{caption}

\usepackage{graphicx}
\usepackage{subfigure}
\usepackage{amsmath}
\usepackage{amssymb}

\title{Knowledge-Embedded Representation Learning \\ for Fine-Grained Image Recognition}

\author{
Tianshui Chen$^1$, 
Liang Lin$^{1,2}$\thanks{\footnotesize Corresponding author is Liang Lin (Email: linliang@ieee.org). This work was supported by the National Natural Science Foundation of China under Grant  61622214, the Science and Technology Planning Project of Guangdong Province under Grant 2017B010116001, and Guangdong Natural Science Foundation Project for Research Teams under Grant 2017A030312006.}, 
Riquan Chen$^1$, 
Yang Wu$^1$
{\normalfont and}  Xiaonan Luo$^3$
\\ 
$^1$ Sun Yat-sen University, China \\
$^2$ SenseTime Research, China\\
$^3$ Guilin University of Electronic Technology, China  \\
tianshuichen@gmail.com,
linliang@ieee.org,\\
sysucrq@gmail.com,
wuyoung567@gmail.com,
luoxn@guet.edu.cn
}
\begin{document}

\maketitle

\begin{abstract}
Humans can naturally understand an image in depth with the aid of rich knowledge accumulated from daily lives or professions. For example, to achieve fine-grained image recognition (e.g., categorizing hundreds of subordinate categories of birds) usually requires a comprehensive visual concept organization including category labels and part-level attributes. In this work, we investigate how to unify rich professional knowledge with deep neural network architectures and propose a Knowledge-Embedded Representation Learning (KERL) framework for handling the problem of fine-grained image recognition. Specifically, we organize the rich visual concepts in the form of knowledge graph and employ a Gated Graph Neural Network to propagate node message through the graph for generating the knowledge representation. By introducing a novel gated mechanism, our KERL framework incorporates this knowledge representation into the discriminative image feature learning, i.e., implicitly associating the specific attributes with the feature maps. Compared with existing methods of fine-grained image classification, our KERL framework has several appealing properties: i) The embedded high-level knowledge enhances the feature representation, thus facilitating distinguishing the subtle differences among subordinate categories. ii) Our framework can learn feature maps with a meaningful configuration that the highlighted regions finely accord with the nodes (specific attributes) of the knowledge graph. Extensive experiments on the widely used Caltech-UCSD bird dataset demonstrate the superiority of our KERL framework over existing state-of-the-art methods.
\end{abstract}

\section{Introduction}

\begin{figure}[!t]
   \centering
   \includegraphics[width=0.95\linewidth]{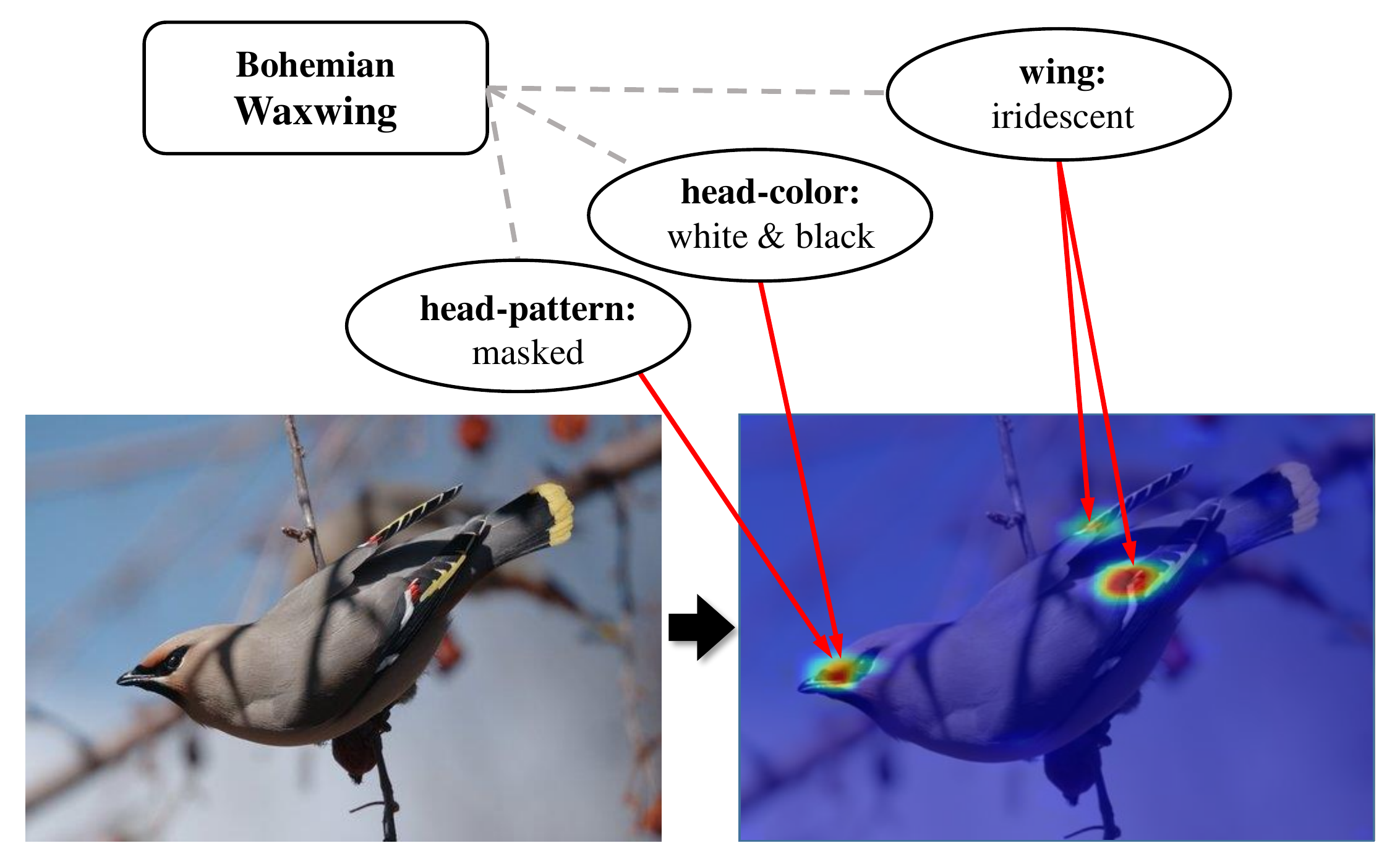} 
   \caption{An example of how the professional knowledge aids fine-grained image understanding. Our proposed framework is capable of associating the specific attributes with the image feature representation.}
   \vspace{-12pt}
   \label{fig:knowledge}
\end{figure}
Humans perform object recognition task based on not only the object appearance but also the knowledge acquired from daily lives or professions. Usually, this knowledge refers to a comprehensive visual concept organization including category labels and their attributes. It is extremely beneficial to fine-grained image classification as attributes are always key to distinguish different subordinate categories. For example, we might know from a book that a bird of category ``Bohemian Waxwing'' has a masked head with color black and white and wings with iridescent feathers. With this knowledge, to recognize the category ``Bohemian Waxwing'' given a bird image, we might first recall the knowledge, attend to the corresponding parts to see whether it possesses these attributes, and then perform reasoning. Figure \ref{fig:knowledge} illustrates an example of how the professional knowledge aids fine-grained image recognition.

Conventional approaches for fine-grained image classification usually neglect this knowledge and merely rely on low-level image cues for recognition. These approaches either employ part-based models \cite{zhang2014part} or resort to visual attention networks \cite{liu2016fully} to locate discriminative regions/parts to distinguish subtle differences among different subordinate categories. However, part-based models involve heavy annotations of object parts, preventing them from application to large-scale data, while visual attention networks can only locate the parts/regions roughly due to the lack of supervision or guidance. Recently, \cite{he2017fine} utilize natural language descriptions to help search the informative regions and combine with vision stream for final prediction. This method also integrates high-level information, but it directly models image-language pairs and requires detailed language descriptions for each image (e.g., ten sentences for each image in  \cite{he2017fine}). Different from these methods, we organize knowledge about categories and part-based attributes in the form of knowledge graph and formulate a Knowledge-Embedded Representation Learning (KERL) framework to incorporate the knowledge graph into image feature learning to promote fine-grained image recognition.

\begin{figure*}[htbp]
   \centering
   \includegraphics[width=0.80\linewidth]{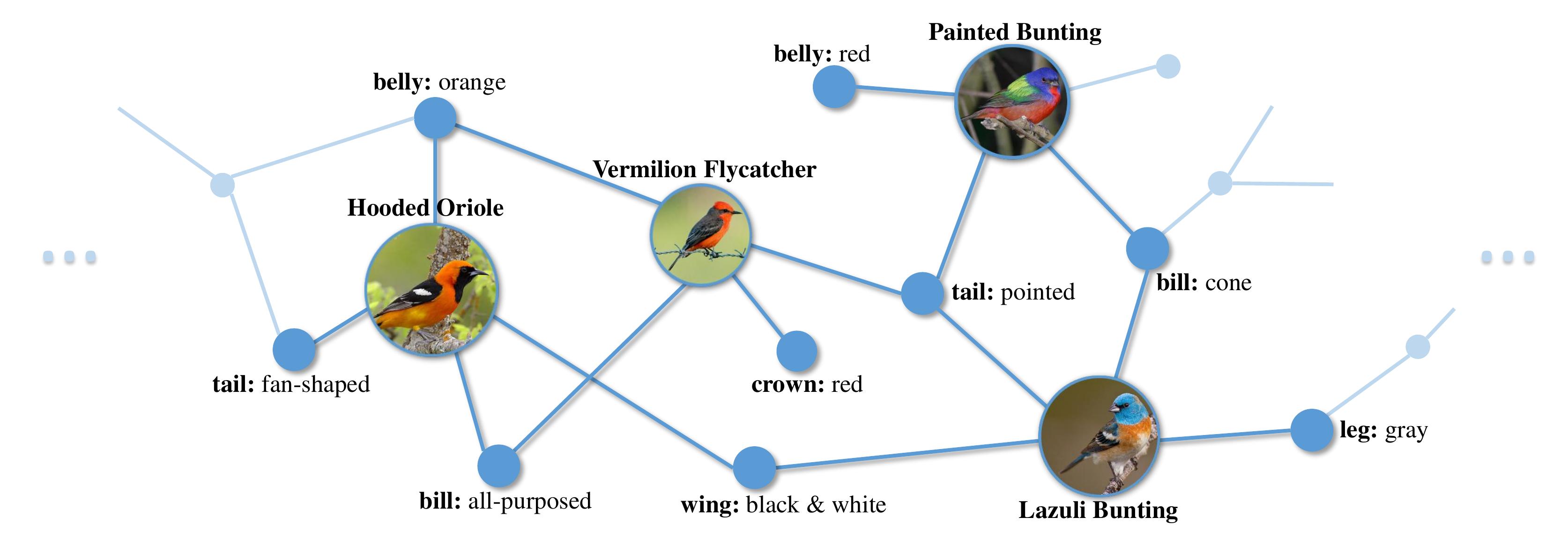} 
   \caption{An example knowledge graph for modeling the category-attribute correlations on the Caltech-UCSD bird dataset.}
   \label{fig:kg}
\end{figure*}

To this end, our proposed KERL framework contains two crucial components: i) a Gated Graph Neural Network (GGNN) \cite{li2015gated} that propagates node message through the graph to generate knowledge representation and ii) a novel gated mechanism that integrates this representation with image feature learning to learn attribute-aware features. Concretely, we first construct a large-scale knowledge graph that relates category labels with part-level attributes as shown in Figure \ref{fig:kg}. By initializing the graph node with information of a given image, our KERL framework might implicitly reason about the discriminative attributes for the image and associate these attributes with feature maps. In this way, our KERL framework can learn feature maps with a meaningful configuration that the highlighted regions finely associate with the relevant attributes in the graph. For example, the learned feature maps of samples from category ``Bohemian Waxwing" always highlight the regions of head and wings, because these regions relate to attributes ``head pattern: masked", ``head color: white \& black" and ``wing: iridescent" that are key to distinguish this category from others. This characteristic also provides insight into why the framework improves performance.

The major contributions of this work are summarized to three-fold: 1) This work formulates a novel Knowledge-Embedded Representation Learning framework that incorporates high-level knowledge graph as extra guidance for image representation learning. To the best of our knowledge, this is the first work to investigate this point. 2) With the guidance of knowledge, our framework can learn attribute-aware feature maps with a meaningful and interpretable configuration that the highlighted regions are finely related to the relevant attributes in the graph, which can also explain performance improvement. 3) We conduct extensive experiments on the widely used Caltech-UCSD bird dataset \cite{wah2011caltech} and demonstrate the superiority of the proposed KERL framework over the leading fine-grained image classification methods.

\section{Related Work}
We review the related work in term of two research streams: fine-grained image classification and knowledge representation.

\subsection{Fine-Grained Image Classification}
With the advancement of deep learning \cite{he2016deep,simonyan2014very}, most works rely on deep Convolutional Neural Networks (CNNs) to learn discriminative features for fine-grained image recognition, which exhibit a notable improvement compared with conventional hand-crafted features \cite{he2017weakly,lin2015bilinear}. To better capture subtle visual difference for fine-grained classification, bilinear models \cite{lin2015bilinear,gao2016compact,kong2017low} is proposed to compute high-order representation that can better model local pairwise feature interactions by two independent sub-networks. Another common approach for distinguishing subtle visual difference among sub-ordinate categories is first locating discriminative regions and then learning appearance model conditioned on these regions \cite{zhang2014part,huang2016part}. However, these methods involve in heavy annotations of object parts, and moreover, manually defined parts may not be optimal for the final recognition. Instead, He et al. \cite{he2017fine} adopt salient region localization techniques \cite{chen2016disc,zhou2016learning} to automatically generate bounding box annotations of the discriminative regions. Recently, visual attention models \cite{mnih2014recurrent,wang2017multi,chen2018recurrent,liu2018crowd} have been intensively proposed to automatically search the informative regions, and some works also apply this technique to fine-grained recognition task \cite{liu2016fully,zheng2017learning,peng2018object}. \cite{liu2016fully} introduce a reinforcement learning framework to adaptively glimpse local discriminative regions and propose a greedy reward strategy to train the framework with image-level annotations. \cite{fu2017look} further introduce a recurrent attention convolutional neural network to recursively learn the attentional regions at multiple scales and region-based feature representation. \cite{liu2017localizing} utilize part-level attribute to guide locating the attentional regions, which is related to ours. However, we organize the category-attribute relationships in the form of knowledge graph and implicitly reason discriminative attributes on the graph rather than using object-attribute pairs directly.




\subsection{Knowledge Representation}
Learning knowledge representation for visual reasoning increasingly receives attention as it benefits various tasks \cite{malisiewicz2009beyond,lao2011random,zhu2014reasoning,lin2017knowledge} in vision community. For instance, \cite{zhu2014reasoning} learn a knowledge base using a Markov Logic Network and employ first-order probabilistic inference to reason the object affordances. These approaches usually involve in hand-crafted features and manually-defined propagation rules, preventing them from end-to-end training. Most recently, a series of efforts are dedicated to adapt neural networks to process graph-structured data \cite{duvenaud2015convolutional,niepert2016learning}. For example, \cite{niepert2016learning} sort the nodes in the graph based on the graph edges to regular sequence and directly feed the node sequence to a standard CNN for feature learning. These methods are tried on small, clean graphs such as molecular datasets \cite{duvenaud2015convolutional} or are used to encode the contextual dependencies for vision tasks \cite{liang2016semantic}.

GGNN \cite{li2015gated} is a fully differentiable recurrent neural network architecture for graph-structured data, which recursively propagates node message to its neighbors to learn node-level features or graph-level representation. Several works have developed a series of graph neural network variants and successfully apply them to various tasks, such as 3DGNN for RGBD semantic segmentation \cite{qi20173d}, model-based GNN for situation recognition \cite{li2017situation}, and GSNN for multi-label image recognition \cite{marino2017more}. Among these works, GSNN \cite{marino2017more} is mostly related to ours in the spirit of GGNN based knowledge graph encoding, but it simply concatenates image and knowledge features for image classification. In contrast, we develop a novel gated mechanism to embed the knowledge representation into image feature learning to enhances the feature representation. Besides, our learned feature maps exhibit insightful configurations that the highlighted regions finely accord with the semantic attributes in the graph, which also provide insight to explain performance improvement.

\section{KERL Framework}
\begin{figure*}[htbp]
   \centering
   \includegraphics[width=0.85\linewidth]{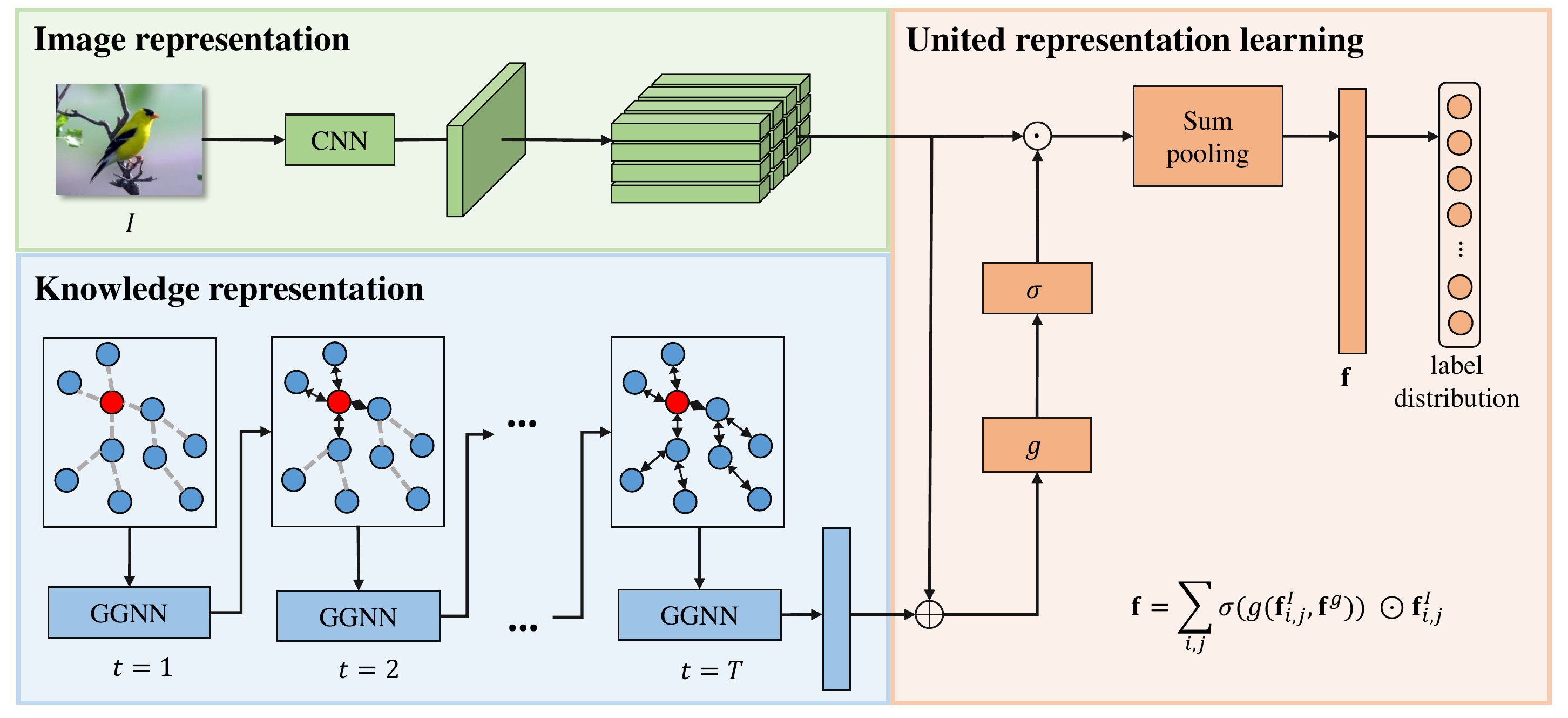} 
   \vspace{-6pt}
   \caption{An overall pipeline of our proposed knowledge-embedded representation learning framework. The framework primally consists of a GGNN that takes the knowledge graph as input and propagates node information through the graph to learn knowledge representation, and a gated mechanism that embeds the representation into the image feature learning to learn attribute-aware features. All components of the framework can be trained in an end-to-end fashion.}
   \vspace{-6pt}
   \label{fig:pipeline}
\end{figure*}

In this section, we first briefly review the GGNN and present the construction of our knowledge graph that relates category labels with their part-level attributes. Then, we introduce our KERL framework in detail, which consists of a GGNN for knowledge representation learning and a gated mechanism to embed knowledge into discriminative image representation learning. An overall pipeline of the framework is illustrated in Figure \ref{fig:pipeline}. 

\subsection{Review of GGNN}
We briefly introduce the GGNN \cite{li2015gated} for completeness. GGNN is recurrent neural network architecture that can learn features for arbitrary graph-structured data by iteratively updating node features. For the propagation process, the input data is represented as a graph $\mathcal{G}=\{\mathbf{V}, \mathbf{A}\}$, in which $\mathbf{V}$ is the node set and $\mathbf{A}$ is the adjacency matrix that denotes the connections among nodes in the graph. For each node $v\in \mathbf{V}$, it has a hidden state $\mathbf{h}_v^t$ at time step $t$, and the hidden state at $t=0$ is initialized by the input feature vector $\mathbf{x}_v$ that depends on the problem in hand. Thus, the basic recurrent process is formulated as 
\begin{equation}
   \begin{split}
    \mathbf{h}_v^0=&{}\mathbf{x}_v\\
    \mathbf{a}_v^t=&{}\mathbf{A}_v^\top[\mathbf{h}_1^{t-1} \dots \mathbf{h}_{|\mathbf{V}|}^{t-1}]^\top+\mathbf{b}\\
    \mathbf{z}_v^t=&{}\sigma(\mathbf{W}^z{\mathbf{a}_v^t}+\mathbf{U}^z{\mathbf{h}_v^{t-1}}) \\
    \mathbf{r}_v^t=&{}\sigma(\mathbf{W}^r{\mathbf{a}_v^t}+\mathbf{U}^r{\mathbf{h}_v^{t-1}}) \\
    \widetilde{\mathbf{h}_v^t}=&{}\tanh\left(\mathbf{W}{\mathbf{a}_v^t}+\mathbf{U}({\mathbf{r}_v^t}\odot{\mathbf{h}_v^{t-1}})\right) \\
    \mathbf{h}_v^t=&{}(1-{\mathbf{z}_v^t}) \odot{\mathbf{h}_v^{t-1}}+{\mathbf{z}_v^t}\odot{\widetilde{\mathbf{h}_v^t}}
   \end{split}
   \label{eq:ggnn}
\end{equation}
where $\mathbf{A}_v$ is a sub-matrix of $\mathbf{A}$ denoting the connections of node $v$ with its neighbors. $\sigma$ and $\tanh$ are the logistic sigmoid and hyperbolic tangent functions, respectively, and $\odot$ denotes the element-wise multiplication operation. The propagation process is repeated until a fixed iteration $T$, and we can obtain the final hidden states $\{\mathbf{h}_1^T, \mathbf{h}_2^T, \dots, \mathbf{h}_{|\mathbf{V}|}^T\}$. For notation simplification, we denote the computation process of equation (\ref{eq:ggnn}) as $\mathbf{h}_v^t=\mathrm{GGNN}(\mathbf{h}_1^{t-1},\dots,\mathbf{h}_{|\mathbf{V}|}^{t-1};\mathbf{A}_v)$.

\subsection{Knowledge Graph Construction}
The knowledge graph refers to an organization of a repository of visual concepts including category labels and part-level attributes, with nodes representing the visual concepts and edges representing their correlations. The graph is constructed based on the attribute annotations of the training samples. An example knowledge graph for the Caltech-UCSD bird dataset \cite{wah2011caltech} is presented in Figure \ref{fig:kg}.


\noindent\textbf{Visual concepts. }A visual concept refers to either a category label or an attribute. The attribute is an intermediate semantic representation of objects, and usually, it is key to distinguish two subordinate categories. Given a dataset that covers $C$ object categories and $A$ attributes, the graph has a node set $\mathbf{V}$ with $C+A$ elements.

\noindent\textbf{Correlation. }The correlation between a category label and an attribute indicates whether this category possesses the corresponding attribute. However, for the fine-grained task, it is common that merely some instances of a category possess a specific attribute. For example, for a specific category, it is possible that one instance has a certain attribute, but another instance does not have. Thus, such category/attribute correlation is uncertain. Fortunately, we can assign an attribute/object instance pair with a score that denotes how likely this instance has the attribute. Then, we can sum up the scores of attribute/object instance pairs for all instances belonging to a specific category and obtain a score to denote the confidence that this category has the attribute. All the scores are linearly normalized to $[0, 1]$ to achieve a $C \times A$ matrix $\mathbf{S}$. Note that no connection exists between two object category nodes or between two attribute nodes; thus the complete adjacency matrix can be expressed as 
\begin{equation}
\mathbf{A}_c=   \left[\begin{matrix} 
      \mathbf{0}_{C \times C} & \mathbf{S} \\
      \mathbf{0}_{A \times C} &  \mathbf{0}_{A \times A} \\
   \end{matrix}\right],
\end{equation}
where $\mathbf{0}_{W \times H}$ denotes a zero matrix of size $W \times H$. In this way, we can construct a knowledge graph $\mathcal{G}=\{\mathbf{V}, \mathbf{A}_c\}$. 

\subsection{Knowledge Representation Learning}
After building the knowledge graph, we employ the GGNN to propagate node message through the graph and compute a feature vector for each node. All the feature vectors are then concatenated to generate the final representation for the knowledge graph. 

We initialize the node referring to category label $i$ with a score $s_i$ that represents the confidence of this category being presented in the given image, and the node referring to each attribute with zero vector. The score vector $\mathbf{s}=\{s_0, s_1, \dots, s_{C-1}\}$ for all categories is estimated by a pre-trained classier that will be introduced in detail in section \ref{sec:exp}. Thus, the input feature for each node can be represented as


\begin{equation}
\mathbf{x}_v=
\begin{cases}
[s_i, \mathbf{0}_{n-1}]& \text{if node $v$ refers to category $i$}\\
[\mathbf{0}_n]& \text{if node $v$ refers to an attribute}
\end{cases},
\end{equation}
where $\mathbf{0}_n$ is a zero vector with dimension $n$. As discussed above, messages of all nodes are propagated to each other during the propagation process. With the computational process of Equation \ref{eq:ggnn}, $\mathbf{A}_c$ are used to propagate message from a certain node to its neighbors, and we use matrix $\mathbf{A}_c^\top$ for reverse message propagation. Thus, the adjacency matrix is $\mathbf{A}=[\mathbf{A}_c \quad \mathbf{A}_c^\top]$.

For each node $v$, its hidden state is initialized using $\mathbf{x}_v$, and at timestep $t$, the hidden state $\mathbf{h}_v^t$ is updated using the propagation process as Equation (\ref{eq:ggnn}),  expressed as

\begin{equation}
   \begin{split}
    \mathbf{h}_v^0=&{}\mathbf{x}_v \\
    \mathbf{h}_v^t=&{}\mathrm{GGNN}(\mathbf{h}_1^{t-1},\dots,\mathbf{h}_{|\mathbf{V}|}^{t-1};\mathbf{A}_v)
   \end{split}.
   \label{eq:upate}
\end{equation}
At each iteration, the hidden state of each node is determined by its history state and the messages sent by its neighbors. In this way, each node can aggregate information from its neighbors and simultaneously transfer its message to its neighbors. This process is shown in Figure \ref{fig:pipeline}. After $T$ iterations, the message of each node has propagated through the graph, and we can get the final hidden state for all nodes in the graph, i.e., $\{\mathbf{h}_1^{T},\mathbf{h}_2^{T},\dots,\mathbf{h}_{|\mathbf{V}|}^{T}\}$. Similar to \cite{li2015gated}, the node-level feature is computed by
\begin{equation}
\mathbf{o}_v=o(\mathbf{h}_v^T, \mathbf{x}_v), v=1,2,\dots,|\mathbf{V}|,
\end{equation} 
where $o$ is an output network that is implemented by a fully-connected layer. Finally, these features are concatenated to produce the final knowledge representation $\mathbf{f}^g$. 

\subsection{United Representation Learning}
In this part, we introduce the gated mechanism that embeds the knowledge representation to enhance image representation learning.


\noindent\textbf{Image feature extraction. }We start by introducing the image feature extraction. As compact bilinear model \cite{gao2016compact} works well on fine-grained image classification, we straightforwardly apply this model to extract image features. Specifically, given an image, we utilize a fully convolutional network (FCN) to extract feature maps with a size of $W'\times H' \times d$, and a compact bilinear operator to produce feature maps $\textbf{f}^I$. Note that we do not perform sum pooling like \cite{gao2016compact} thus the size of $\textbf{f}^I$ is $W'\times H' \times c$. For fair comparisons with existing works, we employ the convolutional layers of the VGG16-Net to implement the FCN and follow the default setting as \cite{gao2016compact} to set $c$ as 8192.

\cite{gao2016compact} treats all features equally important and simply performs sum pooling to obtain $c$-dimensional features for prediction. In the context of fine-grained image classification, it is crucial to attend to the discriminative regions to capture the subtle difference between different subordinate categories. In addition, the knowledge representation encodes category-attribute correlation and it may capture the discriminative attributes. Thus, we embed this representation into image feature learning to learn feature corresponding to this attributes. Specifically, we introduce a gated mechanism that optionally allows the informative features through while suppressing the non-informative features under the guidance of knowledge, which can be formulated as
\begin{equation}
    \mathbf{f}=\sum_{i,j}\sigma\left(g\left(\mathbf{f}_{i,j}^I, \mathbf{f}^g\right)\right)\odot \mathbf{f}_{i,j}^I,
\end{equation}
where $\mathbf{f}_{i,j}^I$ is the feature vector at location $(i,j)$. $\sigma\left(g\left(\mathbf{f}_{i,j}^I, \mathbf{f}^g\right)\right)$ acts as a gated mechanism that decides which location is more important. $g$ is a neural network that takes the concatenation of $\mathbf{f}_{i,j}^I$ and $\mathbf{f}^g$ as input and outputs a $c$-dimensional real-value vector. It is implemented by two stacked fully connected layers in which the first one is 10752 (8192 + 512$\times$5) to 4096 followed by the hyperbolic tangent function while the second one is 4096 to 8192. The feature vector $\mathbf{f}$ is then fed into a simple fully-connected layer to compute the score vector $\mathbf{s}$ for the given image. 

\section{Experiments}

\subsection{Experiment Settings}
\label{sec:exp}
\begin{figure}[!t]
   \centering
   \includegraphics[width=0.9\linewidth]{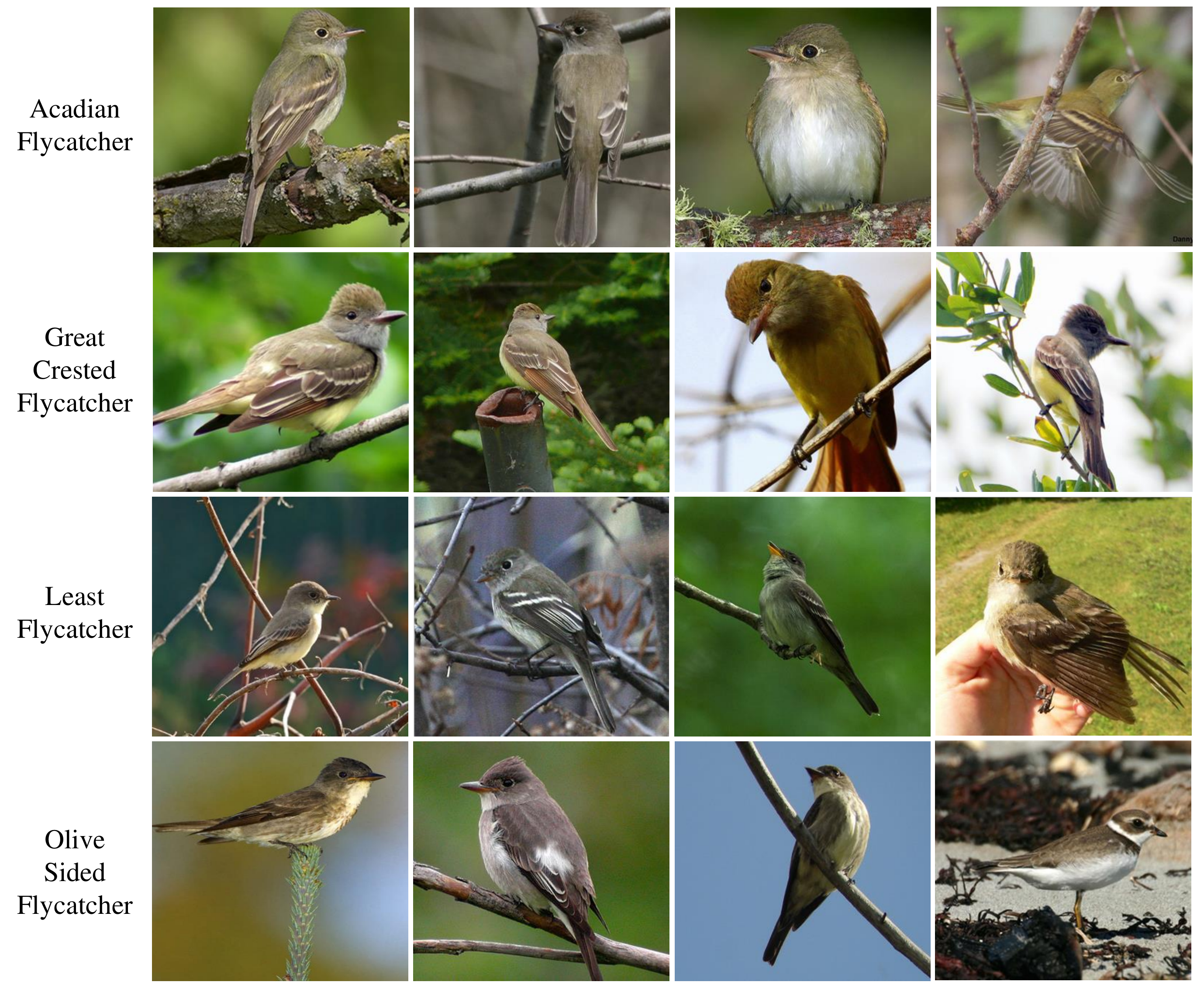} 
   \vspace{-6pt}
   \caption{Samples from the Caltech-UCSD birds dataset. It is extremely difficult to categorize them due to large intra-class variance and small inter-class variance.}
   \vspace{-6pt}
   \label{fig:dataset}
\end{figure}

\noindent\textbf{Datasets. }We evaluate our KERL framework and the competing methods on the Caltech-UCSD bird dataset \cite{wah2011caltech} that is the most widely used benchmark for fine-grained image classification. The dataset covers 200 species of birds, which contains 5,994 images for training and 5,794 for test. Except for the category label, each image is further annotated with 1 bounding box, 15 part key-points, and 312 attributes. As shown in Figure \ref{fig:dataset}, the dataset is extremely challenging because birds from similar species may share very similar visual appearance while birds within the same species undergo drastic changes owing to complex variations in scales, viewpoints, occlusion, and background. In this work, we evaluate the methods in two settings: 1) ``bird in image": the whole image is fed into the model at training and test stages, and 2) ``bird in bbox": the image region at the bounding box is fed into the model at training and test stages.

\noindent\textbf{Implementation details. }For the GGNN, we utilize the compact bilinear model released by work \cite{gao2016compact} to produce the scores to initialize the hidden states. For fair comparisons, the model is implemented with VGG16-Net and trained on the training part of the Caltech-UCSD bird dataset. The dimension of the hidden state is set to 10 and that of the output feature is set to 5. The iteration time $T$ is set to 5. The KERL framework is jointly trained using the cross-entropy loss. All components of the framework are trained with SGD except GGNN that is trained with ADAM following \cite{marino2017more}.

\subsection{Comparison with State-of-the-Art Methods}

In this subsection, we compare our KERL framework with 16 state-of-the-art methods, among which, some use merely image-level labels, and some also use bounding box/parts annotations; thus we also present this information for fair and direct comparisons. The methods are evaluated in both two settings, i.e., ``bird in image" and ``bird in bbox", and the results are reported in Table \ref{table:cub-resule}. 
For the ``bird in bbox" setting, the previous well-performing methods are PN-CNN and SPDA-CNN that achieve the accuracies of 85.4\% and 85.1\% respectively, but they require strong supervision of ground truth part annotations. The accuracy of B-CNN is also up to 85.1\%, but it relies on a very high-dimensional feature representation (250k dimensions). In contrast, the KERL framework requires no ground truth part annotations and utilize a much lower-dimensional feature representation (i.e., 8,192 dimensions), but it achieves an accuracy of 86.6\% that outperforms all previous state-of-the-art methods. 
For the ``bird in image" setting, most existing methods explicitly search discriminative regions and aggregate deep features of these regions for classification. For example, RA-CNN recurrently discovers image regions over three scales and achieves an accuracy of 85.3\%. Besides, CVL combines detailed human-annotated text description for each image with the visual features to further improve the accuracy to 85.6\%. Different from them, the KERL framework learns knowledge representation that encodes category-attribute correlations and incorporates this representation for feature learning. In this way, our method can learn more discriminative attribute-related features, leading to improvement in performance, i.e., 86.3\% in accuracy. Note that AGAL also employs part-level attributes for fine-grained classification, but it achieves accuracies of 85.5\% and 85.4\% in two settings, respectively, much worse than ours. These comparisons well demonstrate the effectiveness of the KERL framework method over existing algorithms. 

\begin{table}[!t]
\centering
\footnotesize
\begin{tabular}{c|c|c|c}
\hline
\centering  Methods  & BA & PA  & Acc. (\%)  \\
\hline
\hline
Part-RCNN \cite{zhang2014part} & $\surd$ & $\surd$ & 76.4 \\
DeepLAC \cite{lin2015deep} & $\surd$ & $\surd$ & 80.3 \\
SPDA-CNN \cite{zhang2016spda} & $\surd$ & $\surd$ & 85.1 \\
PN-CNN \cite{branson2014bird} & $\surd$ & $\surd$ & 85.4 \\
PA-CNN \cite{krause2015fine} & $\surd$ & & 82.8 \\
CB-CNN w/ bbox \cite{gao2016compact} & $\surd$ && 84.6 \\
FCAN w/ bbox \cite{liu2016fully} & $\surd$ &  & 84.7 \\
B-CNN w/ bbox \cite{lin2015bilinear} & $\surd$ & & 85.1 \\
AGAL w/ bbox \cite{liu2017localizing} & $\surd$ & & 85.5 \\
\hline
\textbf{KERL w/ bbox} &$\surd$ & & \textbf{86.6} \\
\textbf{KERL w/ bbox \& w/ HR} &$\surd$ & & \textbf{86.8} \\
\hline
\hline
TLAN \cite{xiao2015application} & & & 77.9 \\
DVAN \cite{zhao2017diversified} & & & 79.0 \\
MG-CNN \cite{wang2015multiple} & & &81.7 \\
B-CNN w/o bbox \cite{lin2015bilinear} & & & 84.1 \\
ST-CNN \cite{jaderberg2015spatial} & & & 84.1 \\
FCAN w/o bbox \cite{liu2016fully} & &  & 84.3 \\
PDFR \cite{zhang2016picking} & & & 84.5 \\
CB-CNN w/o bbox \cite{gao2016compact} &&& 85.0 \\
RA-CNN \cite{fu2017look} & &  & 85.3 \\
AGAL w/o bbox \cite{liu2017localizing} & &  & 85.4 \\
CVL \cite{he2017fine} & & & 85.6 \\
\hline
\textbf{KERL} & & & \textbf{86.3} \\ 
\textbf{KERL w/ HR} & & & \textbf{87.0} \\ 
\hline
\end{tabular}
\caption{Comparisons of our KERL framework with existing state of the arts on the Caltech-UCSD bird dataset. BA and PA denote bounding box annotations and part annotations, respectively, and HR denotes highlighted regions. $\surd$ indicates corresponding annotations are used during training or test.}
\label{table:cub-resule}
\end{table}

Attention-based methods aggregate features of both image and located regions to promote fine-grained classification, and our results reported above merely use image features. Our KERL framework can learn feature maps that highlight the regions related to discriminative attributes, as discussed in section \ref{RV}; thus, we also aggregate features of the highlighted regions to improve performance. Specifically, we sum up the feature values across channels to get a score at each location, and draw a region with a size of a $6 \times 6$ centered at each location. We adopt non-maximum suppression to exclude the seriously overlapped regions and select top three ones. Three corresponding regions with a size of $96 \times 96$ (16$\times$ mapping between the original image and feature map) in the image are cropped, resized to $224 \times 224$ and fed to the VGG16 net to extract feature, respectively. The features are concatenated and fed to a fully-connected layer to compute the score vector, which is further averaged with the results of KERL to achieve the final results. It boosts the accuracies to 86.8\% and 87.0\% in two settings respectively.

\subsection{Contribution of Knowledge Embedding}
Note that our KERL framework employs CB-CNN \cite{gao2016compact} as the baseline. Here, we emphasize the comparison with this baseline method to demonstrate the significance of knowledge embedding knowledge. As shown in Table \ref{table:cub-knowledge}, the CB-CNN achieves accuracies of 84.6\% and 85.0\% in ``bird in bbox" and ``bird in image" settings. By embedding the knowledge representation, the KERL framework boosts the accuracies to 86.6\% and 86.3\%, improving those of the CB-CNN by 2.0\% and 1.3\%, respectively. 

To further clarify the contribution of knowledge guided feature selection, we implement two more baseline methods: self-guided feature learning and feature concatenation.

\noindent\textbf{Comparison with self-guided feature learning. }To better verify the benefit of embedding knowledge for feature learning, we conduct an experiment that removes the GGNN and only feeds the image features to the gated neural network, with other components left unchanged. The comparison results are presented in Table \ref{table:cub-knowledge}. It merely exhibits minor improvement over the baseline CB-CNN as it does not incur additional information but only increasing the complexity of the model. As expected, it performs much worse than ours.

\noindent\textbf{Comparison with feature concatenation. }To validate the benefit of our knowledge embedding method, we further conduct an experiment that incorporates knowledge by simply concatenating the image and graph feature vectors, followed by a fully-connected layer for classification. As shown in Table \ref{table:cub-knowledge}, directly concatenating image and graph features can achieve accuracies of 85.4\% and 85.5\% in the two settings, which is slightly better than the original CB-CNN but still much worse than ours. This indicates our knowledge incorporation method can make better use of knowledge to facilitate fine-grained image classification.

\begin{table}[htbp]
\centering
\begin{tabular}{c|c|c}
\hline
\centering  Methods  & ``bird in bbox" & ``bird in image"  \\
\hline
\hline
CB-CNN & 84.6 & 85.0 \\
self-guided selection & 84.8 & 85.3\\
concatenation & 85.4& 85.5\\
Ours & \textbf{86.6} & \textbf{86.3} \\
\hline
\end{tabular}
\vspace{-6pt}
\caption{Accuracy comparisons (in \%) of our KERL framework, feature concatenation, self-guided feature selection and baseline CB-CNN model on the Caltech-UCSD bird dataset.}
\vspace{-6pt}
\label{table:cub-knowledge}
\end{table}

\begin{figure}[!t]
   \centering
   \includegraphics[width=1.0\linewidth]{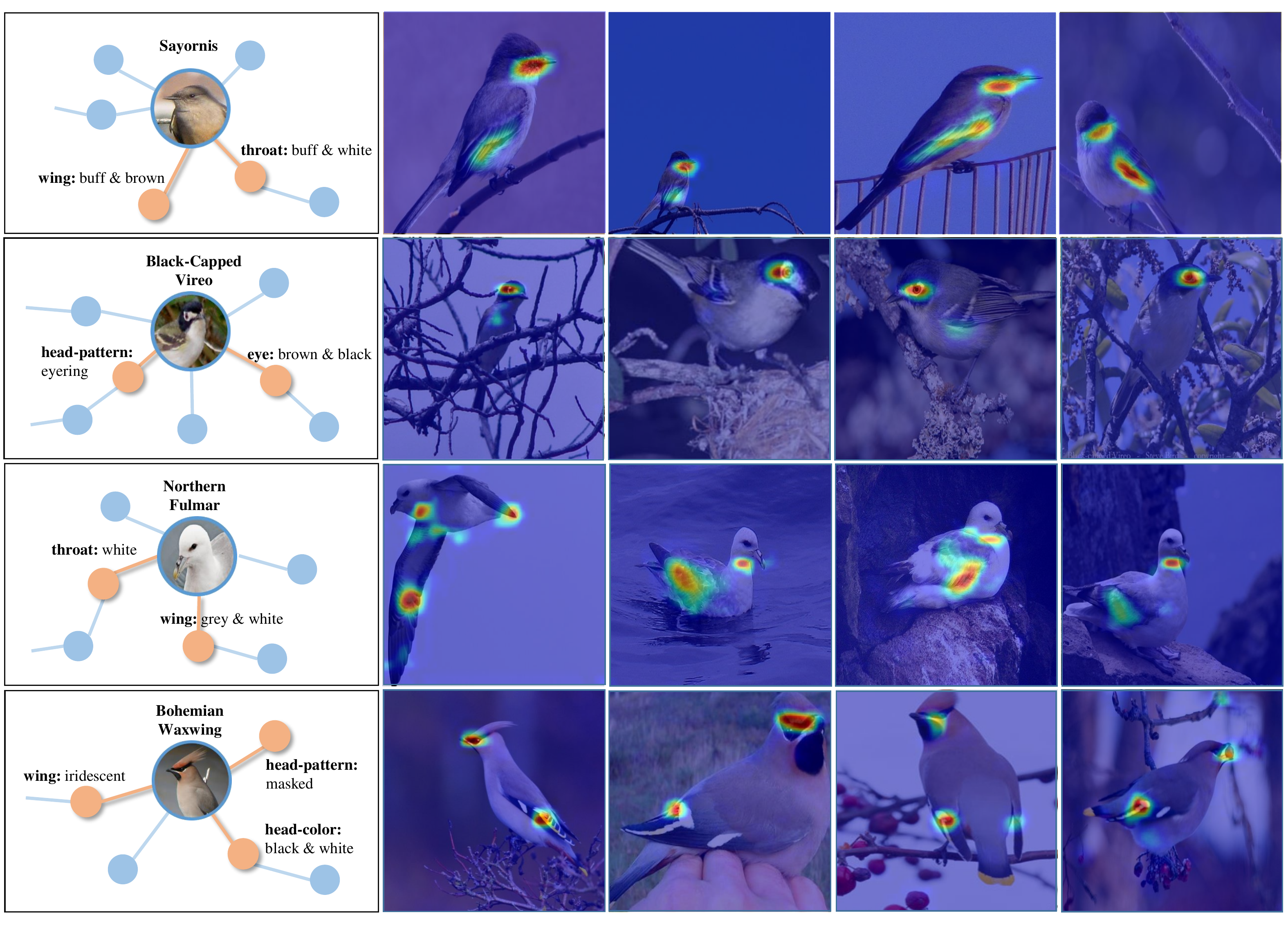} 
   \vspace{-6pt}
   \caption{Visualization of the feature maps learnt by our KERL framework. At each row, we present some samples of a specific category and a sub-graph that denotes the correlations of this category with its attributes. The relevant attributes are highlighted in orange circles.}
   \vspace{-6pt}
   \label{fig:visualization2}
\end{figure}

\subsection{Representation Visualization}
\label{RV}
With knowledge embedding, our KERL framework can learn feature maps with an insightful configuration that the highlighted regions are always related to relevant attributes. Here, we visualize the feature maps before sum pooling to better evaluate this point in Figure \ref{fig:visualization2}. We sum up the feature values across channels at each location and normalize them to $[0, 1]$. At each row, we present the learned feature maps of several samples taken from a specific category and a sub-graph that shows the correlations of this category with its attributes. We find that the highlighted regions for samples of the same category refer to the same semantic parts, and these parts finely accord with the attributes that well distinguish this category from others. Taking the category of ``Sayornis" as example, our KERL framework consistently highlights the regions of throats and wings for all samples, which correspond to two key attributes, i.e., ``throat: buff \& white" and ``wing: buff \& brown" (highlighted with orange circle in Figure \ref{fig:visualization2}). This suggests our KERL framework can learn attribute-aware features that can better capture subtle differences between different subordinate categories. Also, it can provide an explanation for the performance improvement of our framework. 

\begin{figure}[htbp]
   \centering
   \includegraphics[width=0.9\linewidth]{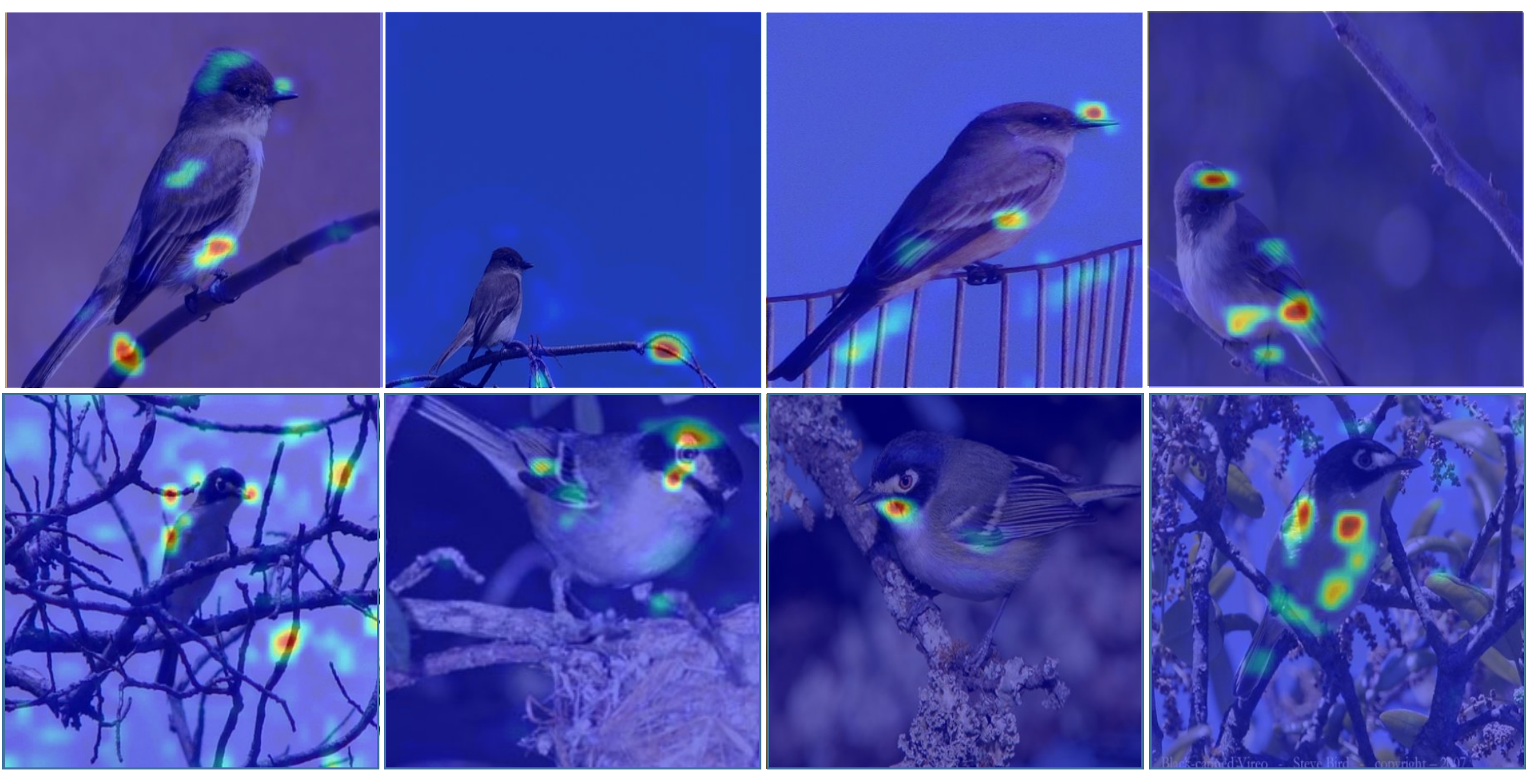} 
   \caption{Visualization of the feature maps generated by the baseline CB-CNN model. The samples are the same with those of the first two categories in Figure \ref{fig:visualization2} for direct comparison.}
   \label{fig:visualization3}
\end{figure}

To clearly verify that it is the knowledge embedding that brings about such appealing characteristic, we further visualize the feature maps generated by the CB-CNN model in Figure \ref{fig:visualization3}. We visualize the samples the same with those of the first two categories in Figure \ref{fig:visualization2} for direct comparison. It is observed that some highlighted regions lie in the background and some scatter over the whole body of the birds.

\section{Conclusion}
In this paper, we propose a novel Knowledge-Embedded Representation Learning (KERL) framework to incorporate knowledge graph as extra guidance for image feature learning. Specifically, the KERL framework consists of a GGNN to learn the graph representation, and a gated neural network to integrate this representation into image feature learning to learn attribute-aware features. Besides, our framework can learn feature maps with an insightful configuration that the highlighted regions are always related to the relevant attributes in the graph, and this can well explain the performance improvement of our KERL framework. Experiments and evaluations conducted on the Caltech-UCSD bird dataset well demonstrate the superiority of our KERL framework over existing state-of-the-art methods. It is an early attempt to embed high-level knowledge into the modern deep network to improve fine-grained image classification, and we hope it can provide a step towards the integration of knowledge and traditional computer vision frameworks.


\small
\bibliographystyle{named}
\bibliography{ijcai18}

\begin{thebibliography}{}

\bibitem[\protect\citeauthoryear{Branson \bgroup \em et al.\egroup
  }{2014}]{branson2014bird}
Steve Branson, Grant Van~Horn, Serge Belongie, and Pietro Perona.
\newblock Bird species categorization using pose normalized deep convolutional
  nets.
\newblock {\em arXiv preprint arXiv:1406.2952}, 2014.

\bibitem[\protect\citeauthoryear{Chen \bgroup \em et al.\egroup
  }{2016}]{chen2016disc}
Tianshui Chen, Liang Lin, Lingbo Liu, Xiaonan Luo, and Xuelong Li.
\newblock Disc: Deep image saliency computing via progressive representation
  learning.
\newblock {\em TNNLS}, 27(6):1135--1149, 2016.

\bibitem[\protect\citeauthoryear{Chen \bgroup \em et al.\egroup
  }{2018}]{chen2018recurrent}
Tianshui Chen, Zhouxia Wang, Guanbin Li, and Liang Lin.
\newblock Recurrent attentional reinforcement learning for multi-label image
  recognition.
\newblock In {\em AAAI}, pages 6730--6737, 2018.

\bibitem[\protect\citeauthoryear{Duvenaud \bgroup \em et al.\egroup
  }{2015}]{duvenaud2015convolutional}
David~K Duvenaud, Dougal Maclaurin, Jorge Iparraguirre, Rafael Bombarell,
  Timothy Hirzel, Al{\'a}n Aspuru-Guzik, and Ryan~P Adams.
\newblock Convolutional networks on graphs for learning molecular fingerprints.
\newblock In {\em NIPS}, pages 2224--2232, 2015.

\bibitem[\protect\citeauthoryear{Fu \bgroup \em et al.\egroup
  }{2017}]{fu2017look}
Jianlong Fu, Heliang Zheng, and Tao Mei.
\newblock Look closer to see better: recurrent attention convolutional neural
  network for fine-grained image recognition.
\newblock In {\em CVPR}, pages 4438--4446, 2017.

\bibitem[\protect\citeauthoryear{Gao \bgroup \em et al.\egroup
  }{2016}]{gao2016compact}
Yang Gao, Oscar Beijbom, Ning Zhang, and Trevor Darrell.
\newblock Compact bilinear pooling.
\newblock In {\em CVPR}, pages 317--326, 2016.

\bibitem[\protect\citeauthoryear{He and Peng}{2017a}]{he2017fine}
Xiangteng He and Yuxin Peng.
\newblock Fine-graind image classification via combining vision and language.
\newblock {\em CVPR}, pages 5994--6002, 2017.

\bibitem[\protect\citeauthoryear{He and Peng}{2017b}]{he2017weakly}
Xiangteng He and Yuxin Peng.
\newblock Weakly supervised learning of part selection model with spatial
  constraints for fine-grained image classification.
\newblock In {\em AAAI}, pages 4075--4081, 2017.

\bibitem[\protect\citeauthoryear{He \bgroup \em et al.\egroup
  }{2016}]{he2016deep}
Kaiming He, Xiangyu Zhang, Shaoqing Ren, and Jian Sun.
\newblock Deep residual learning for image recognition.
\newblock In {\em CVPR}, pages 770--778, 2016.

\bibitem[\protect\citeauthoryear{Huang \bgroup \em et al.\egroup
  }{2016}]{huang2016part}
Shaoli Huang, Zhe Xu, Dacheng Tao, and Ya~Zhang.
\newblock Part-stacked cnn for fine-grained visual categorization.
\newblock In {\em CVPR}, pages 1173--1182, 2016.

\bibitem[\protect\citeauthoryear{Jaderberg \bgroup \em et al.\egroup
  }{2015}]{jaderberg2015spatial}
Max Jaderberg, Karen Simonyan, Andrew Zisserman, et~al.
\newblock Spatial transformer networks.
\newblock In {\em NIPS}, pages 2017--2025, 2015.

\bibitem[\protect\citeauthoryear{Kong and Fowlkes}{2017}]{kong2017low}
Shu Kong and Charless Fowlkes.
\newblock Low-rank bilinear pooling for fine-grained classification.
\newblock In {\em CVPR}, pages 7025--7034, 2017.

\bibitem[\protect\citeauthoryear{Krause \bgroup \em et al.\egroup
  }{2015}]{krause2015fine}
Jonathan Krause, Hailin Jin, Jianchao Yang, and Li~Fei-Fei.
\newblock Fine-grained recognition without part annotations.
\newblock In {\em CVPR}, pages 5546--5555, 2015.

\bibitem[\protect\citeauthoryear{Lao \bgroup \em et al.\egroup
  }{2011}]{lao2011random}
Ni~Lao, Tom Mitchell, and William~W Cohen.
\newblock Random walk inference and learning in a large scale knowledge base.
\newblock In {\em EMNLP}, pages 529--539, 2011.

\bibitem[\protect\citeauthoryear{Li \bgroup \em et al.\egroup
  }{2015}]{li2015gated}
Yujia Li, Daniel Tarlow, Marc Brockschmidt, and Richard Zemel.
\newblock Gated graph sequence neural networks.
\newblock {\em arXiv preprint arXiv:1511.05493}, 2015.

\bibitem[\protect\citeauthoryear{Li \bgroup \em et al.\egroup
  }{2017}]{li2017situation}
Ruiyu Li, Makarand Tapaswi, Renjie Liao, Jiaya Jia, Raquel Urtasun, and Sanja
  Fidler.
\newblock Situation recognition with graph neural networks.
\newblock In {\em CVPR}, pages 4173--4182, 2017.

\bibitem[\protect\citeauthoryear{Liang \bgroup \em et al.\egroup
  }{2016}]{liang2016semantic}
Xiaodan Liang, Xiaohui Shen, Jiashi Feng, Liang Lin, and Shuicheng Yan.
\newblock Semantic object parsing with graph lstm.
\newblock In {\em ECCV}, pages 125--143, 2016.

\bibitem[\protect\citeauthoryear{Lin \bgroup \em et al.\egroup
  }{2015a}]{lin2015deep}
Di~Lin, Xiaoyong Shen, Cewu Lu, and Jiaya Jia.
\newblock Deep lac: Deep localization, alignment and classification for
  fine-grained recognition.
\newblock In {\em CVPR}, pages 1666--1674, 2015.

\bibitem[\protect\citeauthoryear{Lin \bgroup \em et al.\egroup
  }{2015b}]{lin2015bilinear}
Tsung-Yu Lin, Aruni RoyChowdhury, and Subhransu Maji.
\newblock Bilinear cnn models for fine-grained visual recognition.
\newblock In {\em ICCV}, pages 1449--1457, 2015.

\bibitem[\protect\citeauthoryear{Lin \bgroup \em et al.\egroup
  }{2017}]{lin2017knowledge}
Liang Lin, Lili Huang, Tianshui Chen, Yukang Gan, and Hui Cheng.
\newblock Knowledge-guided recurrent neural network learning for task-oriented
  action prediction.
\newblock In {\em ICME}, pages 625--630, 2017.

\bibitem[\protect\citeauthoryear{Liu \bgroup \em et al.\egroup
  }{2016}]{liu2016fully}
Xiao Liu, Tian Xia, Jiang Wang, and Yuanqing Lin.
\newblock Fully convolutional attention localization networks: Efficient
  attention localization for fine-grained recognition.
\newblock {\em arXiv preprint arXiv:1603.06765}, 2016.

\bibitem[\protect\citeauthoryear{Liu \bgroup \em et al.\egroup
  }{2017}]{liu2017localizing}
Xiao Liu, Jiang Wang, Shilei Wen, Errui Ding, and Yuanqing Lin.
\newblock Localizing by describing: Attribute-guided attention localization for
  fine-grained recognition.
\newblock In {\em AAAI}, pages 4190--4196, 2017.

\bibitem[\protect\citeauthoryear{Liu \bgroup \em et al.\egroup
  }{2018}]{liu2018crowd}
Lingbo Liu, Hongjun Wang, Guanbin Li, Wanli Ouyang, and Liang Lin.
\newblock Crowd counting using deep recurrent spatial-aware network.
\newblock In {\em IJCAI}, 2018.

\bibitem[\protect\citeauthoryear{Malisiewicz and
  Efros}{2009}]{malisiewicz2009beyond}
Tomasz Malisiewicz and Alyosha Efros.
\newblock Beyond categories: The visual memex model for reasoning about object
  relationships.
\newblock In {\em NIPS}, pages 1222--1230, 2009.

\bibitem[\protect\citeauthoryear{Marino \bgroup \em et al.\egroup
  }{2017}]{marino2017more}
Kenneth Marino, Ruslan Salakhutdinov, and Abhinav Gupta.
\newblock The more you know: Using knowledge graphs for image classification.
\newblock In {\em CVPR}, pages 2673--2681, 2017.

\bibitem[\protect\citeauthoryear{Mnih \bgroup \em et al.\egroup
  }{2014}]{mnih2014recurrent}
Volodymyr Mnih, Nicolas Heess, Alex Graves, et~al.
\newblock Recurrent models of visual attention.
\newblock In {\em NIPS}, pages 2204--2212, 2014.

\bibitem[\protect\citeauthoryear{Niepert \bgroup \em et al.\egroup
  }{2016}]{niepert2016learning}
Mathias Niepert, Mohamed Ahmed, and Konstantin Kutzkov.
\newblock Learning convolutional neural networks for graphs.
\newblock In {\em ICML}, pages 2014--2023, 2016.

\bibitem[\protect\citeauthoryear{Peng \bgroup \em et al.\egroup
  }{2018}]{peng2018object}
Yuxin Peng, Xiangteng He, and Junjie Zhao.
\newblock Object-part attention model for fine-grained image classification.
\newblock {\em TIP}, 27(3):1487--1500, 2018.

\bibitem[\protect\citeauthoryear{Qi \bgroup \em et al.\egroup
  }{2017}]{qi20173d}
Xiaojuan Qi, Renjie Liao, Jiaya Jia, Sanja Fidler, and Raquel Urtasun.
\newblock 3d graph neural networks for rgbd semantic segmentation.
\newblock In {\em CVPR}, pages 5199--5208, 2017.

\bibitem[\protect\citeauthoryear{Simonyan and
  Zisserman}{2014}]{simonyan2014very}
Karen Simonyan and Andrew Zisserman.
\newblock Very deep convolutional networks for large-scale image recognition.
\newblock {\em arXiv preprint arXiv:1409.1556}, 2014.

\bibitem[\protect\citeauthoryear{Wah \bgroup \em et al.\egroup
  }{2011}]{wah2011caltech}
Catherine Wah, Steve Branson, Peter Welinder, Pietro Perona, and Serge
  Belongie.
\newblock The caltech-ucsd birds-200-2011 dataset.
\newblock 2011.

\bibitem[\protect\citeauthoryear{Wang \bgroup \em et al.\egroup
  }{2015}]{wang2015multiple}
Dequan Wang, Zhiqiang Shen, Jie Shao, Wei Zhang, Xiangyang Xue, and Zheng
  Zhang.
\newblock Multiple granularity descriptors for fine-grained categorization.
\newblock In {\em ICCV}, pages 2399--2406, 2015.

\bibitem[\protect\citeauthoryear{Wang \bgroup \em et al.\egroup
  }{2017}]{wang2017multi}
Zhouxia Wang, Tianshui Chen, Guanbin Li, Ruijia Xu, and Liang Lin.
\newblock Multi-label image recognition by recurrently discovering attentional
  regions.
\newblock In {\em ICCV}, pages 464--472, 2017.

\bibitem[\protect\citeauthoryear{Xiao \bgroup \em et al.\egroup
  }{2015}]{xiao2015application}
Tianjun Xiao, Yichong Xu, Kuiyuan Yang, Jiaxing Zhang, Yuxin Peng, and Zheng
  Zhang.
\newblock The application of two-level attention models in deep convolutional
  neural network for fine-grained image classification.
\newblock In {\em CVPR}, pages 842--850, 2015.

\bibitem[\protect\citeauthoryear{Zhang \bgroup \em et al.\egroup
  }{2014}]{zhang2014part}
Ning Zhang, Jeff Donahue, Ross Girshick, and Trevor Darrell.
\newblock Part-based r-cnns for fine-grained category detection.
\newblock In {\em ECCV}, pages 834--849, 2014.

\bibitem[\protect\citeauthoryear{Zhang \bgroup \em et al.\egroup
  }{2016a}]{zhang2016spda}
Han Zhang, Tao Xu, Mohamed Elhoseiny, Xiaolei Huang, Shaoting Zhang, Ahmed
  Elgammal, and Dimitris Metaxas.
\newblock Spda-cnn: Unifying semantic part detection and abstraction for
  fine-grained recognition.
\newblock In {\em CVPR}, pages 1143--1152, 2016.

\bibitem[\protect\citeauthoryear{Zhang \bgroup \em et al.\egroup
  }{2016b}]{zhang2016picking}
Xiaopeng Zhang, Hongkai Xiong, Wengang Zhou, Weiyao Lin, and Qi~Tian.
\newblock Picking deep filter responses for fine-grained image recognition.
\newblock In {\em CVPR}, pages 1134--1142, 2016.

\bibitem[\protect\citeauthoryear{Zhao \bgroup \em et al.\egroup
  }{2017}]{zhao2017diversified}
Bo~Zhao, Xiao Wu, Jiashi Feng, Qiang Peng, and Shuicheng Yan.
\newblock Diversified visual attention networks for fine-grained object
  classification.
\newblock {\em TMM}, pages 1245--1256, 2017.

\bibitem[\protect\citeauthoryear{Zheng \bgroup \em et al.\egroup
  }{2017}]{zheng2017learning}
Heliang Zheng, Jianlong Fu, Tao Mei, and Jiebo Luo.
\newblock Learning multi-attention convolutional neural network for
  fine-grained image recognition.
\newblock In {\em CVPR}, pages 396--404, 2017.

\bibitem[\protect\citeauthoryear{Zhou \bgroup \em et al.\egroup
  }{2016}]{zhou2016learning}
Bolei Zhou, Aditya Khosla, Agata Lapedriza, Aude Oliva, and Antonio Torralba.
\newblock Learning deep features for discriminative localization.
\newblock In {\em CVPR}, pages 2921--2929, 2016.

\bibitem[\protect\citeauthoryear{Zhu \bgroup \em et al.\egroup
  }{2014}]{zhu2014reasoning}
Yuke Zhu, Alireza Fathi, and Li~Fei-Fei.
\newblock Reasoning about object affordances in a knowledge base
  representation.
\newblock In {\em ECCV}, pages 408--424, 2014.

\end{thebibliography}

\end{document}